\documentclass[10pt,twocolumn,letterpaper]{article}

\usepackage{cvpr}
\usepackage{times}
\usepackage{epsfig}
\usepackage{graphicx}
\usepackage{amsmath}
\usepackage{amssymb}

\usepackage{algorithm,algpseudocode}
\usepackage{array}
\usepackage{color}
\usepackage{subfiles}
\usepackage{setspace}
\usepackage{soul} % for revision
\usepackage{multirow}
\usepackage{caption}
\usepackage{subcaption}
\usepackage[table]{xcolor}
\usepackage{authblk}

\usepackage[pagebackref=true,breaklinks=true,letterpaper=true,colorlinks,bookmarks=false]{hyperref}

\cvprfinalcopy % *** Uncomment this line for the final submission

\def\cvprPaperID{974} % *** Enter the CVPR Paper ID here

\ifcvprfinal\fi
\begin{document}

\makeatletter
\renewcommand*{\@fnsymbol}[1]{\dag}
\makeatother

%%%%%%%%% TITLE
\title{Weakly Supervised Instance Segmentation using Class Peak Response}

\author[1]{Yanzhao Zhou$^\dag$}
\author[1]{Yi Zhu}
\author[1]{Qixiang Ye}
\author[2]{Qiang Qiu}
\author[1]{Jianbin Jiao\thanks{Corresponding Authors}}

\affil[1]{University of Chinese Academy of Sciences}
\affil[2]{Duke University
\authorcr\small \{zhouyanzhao215, zhuyi215\}@mails.ucas.ac.cn, \{qxye, jiaojb\}@ucas.ac.cn, qiang.qiu@duke.edu}

\renewcommand\Authands{ and }

\maketitle
%\thispagestyle{empty}

%%%%%%%%% ABSTRACT
\begin{abstract}
    Weakly supervised instance segmentation with image-level labels, instead of expensive pixel-level masks, remains unexplored. In this paper, we tackle this challenging problem by exploiting class peak responses to enable a classification network for instance mask extraction. With image labels supervision only, CNN classifiers in a fully convolutional manner can produce class response maps, which specify classification confidence at each image location. We observed that local maximums, \ie, peaks, in a class response map typically correspond to strong visual cues residing inside each instance. Motivated by this, we first design a process to stimulate peaks to emerge from a class response map. The emerged peaks are then back-propagated and effectively mapped to highly informative regions of each object instance, such as instance boundaries. We refer to the above maps generated from class peak responses as Peak Response Maps (PRMs). PRMs provide a fine-detailed instance-level representation, which allows instance masks to be extracted even with some off-the-shelf methods. To the best of our knowledge, we for the first time report results for the challenging image-level supervised instance segmentation task. Extensive experiments show that our method also boosts weakly supervised pointwise localization as well as semantic segmentation performance, and reports state-of-the-art results on popular benchmarks, including PASCAL VOC 2012 and MS COCO.
    \footnote{Source code is publicly available at \href{http://yzhou.work/PRM/}{yzhou.work/PRM}}
\end{abstract}

%%%%%%%%% INTRODUCTION
\section{Introduction}

\begin{figure}[!tp]
    \includegraphics[width=1\linewidth]{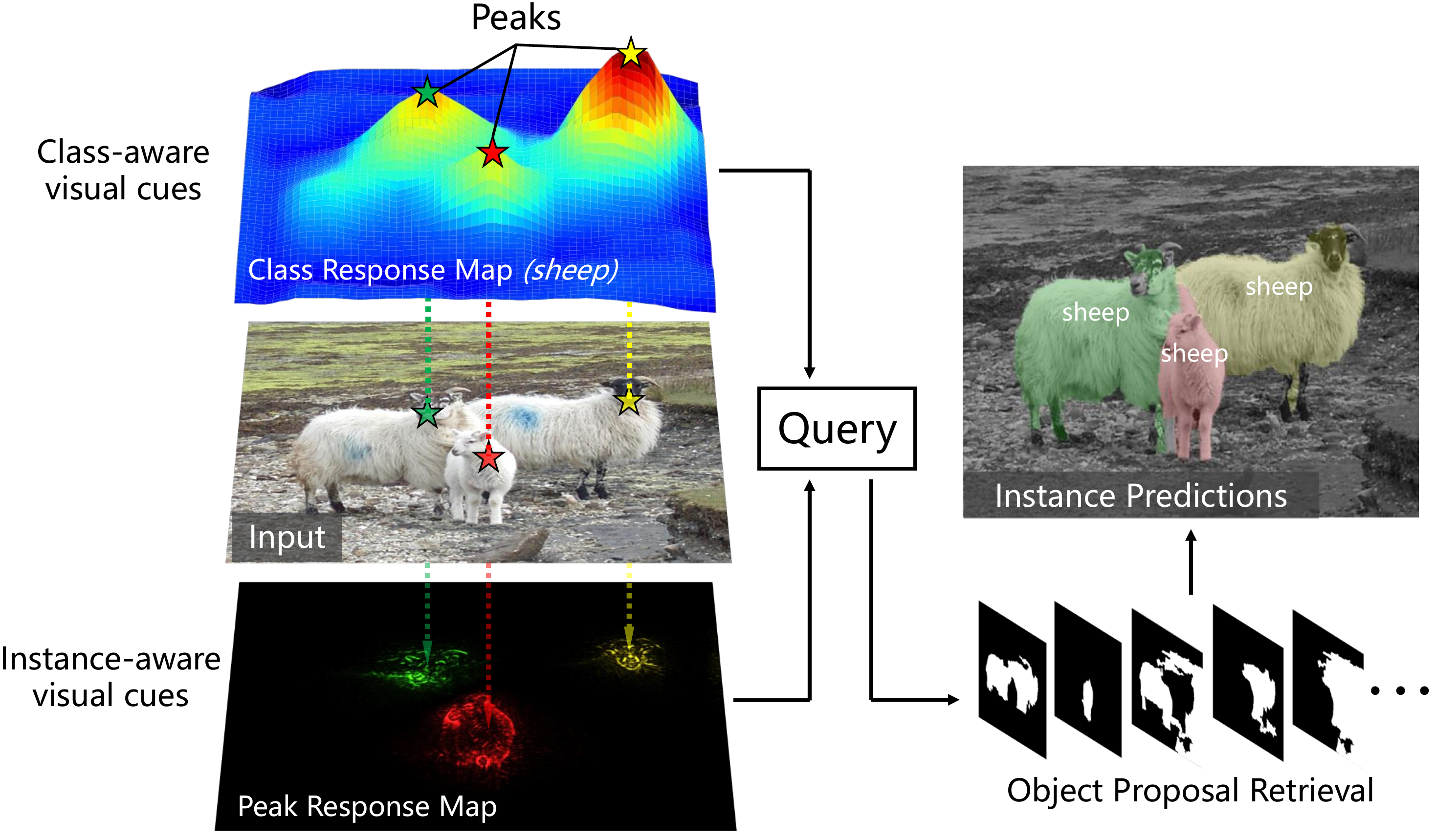}
    \caption{Class peak responses correspond to strong visual cues residing inside each respective instance. 
    Those peaks can be back-propagated and effectively mapped to highly informative regions of each object, which allow instance masks to be extracted. Best viewed in color.}
    \label{fig:cover}
    \vspace{-0.8em}
\end{figure}

Most contemporary methods of semantic segmentation rely on large-scale dense annotations for training deep models; however, annotating pixel-level masks is expensive and labor-intensive \cite{lin2014microsoft}. 
In contrast, image-level annotations, \ie, presence or absence of object categories in an image, are much cheaper and easier to define. 
This motivates the development of weakly supervised semantic segmentation methods, which use image labels to learn convolutional neural networks (CNNs) for class-aware segmentation. 

Most existing weakly supervised semantic segmentation methods consider convolutional filters in CNN as object detectors and aggregate the deep feature maps to extract class-aware visual evidence \cite{zhou2016cnnlocalization, zhang2016EB}. Typically, pre-trained classification networks are first converted to fully convolutional networks (FCNs) to produce class response maps in a single forward pass. 
Such class response maps indicate essential image regions used by the network to identify an image class; however, cannot distinguish different object instances from the same category. Therefore, existing weakly supervised semantic segmentation methods cannot be simply generalized to instance-level semantic segmentation \cite{li2017fully, he2017mask}, which aims to detect all objects in an image as well as predicting precise masks for each instance. 

In this paper, we explore the challenging problem of training CNNs with image-level weak supervision for instance-level semantic segmentation (instance segmentation for short). Specifically, we propose to exploit peaks in a class response map to enable a classification network, \eg, VGGNet, ResNet, for instance mask extraction.

Local maximums, \ie, peaks, in a class response map typically correspond to strong visual cues residing inside an instance, Fig.~\ref{fig:cover}.
Motivated by such observation, we first design a process to stimulate, during the training stage, peaks to emerge from a class response map. At the inference stage, the emerged peaks are back-propagated and effectively mapped to highly informative regions of each object instance, such as instance boundaries. The above maps generated from class peak responses are referred to as Peak Response Maps (PRMs). As shown in Fig.~\ref{fig:cover}, PRMs serve as an instance-level representation, which specifies both spatial layouts and fine-detailed boundaries of each object; thus allows instance masks to be extracted even with some off-the-shelf methods \cite{boykov2001interactive, uijlings2013selective, jordi2015mcg}.

Compared with many fully supervised approaches that typically use complex frameworks including conditional random fields (CRF) \cite{zhang2015monocular,zhang2016instance},  recurrent neural networks (RNN) \cite{ren2016end,romera2016recurrent}, or template matching \cite{uhrig2016pixel}, to handle instance extraction; our approach is simple yet effective. It is compatible with any modern network architectures and can be trained using standard classification settings, \eg, image class labels and cross entropy loss, with negligible computational overhead.
Thanks to its training efficiency, our method is well suited for application to large-scale data.

To summarize, the main contributions of this paper are:
\begin{itemize}
    \item We observe that peaks in class response maps typically correspond to strong visual cues residing inside each respective instance, and such simple observation leads to an effective weakly supervised instance segmentation technique. 

    \item We propose to exploit class peak responses to enable a classification network for instance mask extraction. We first stimulate peaks to emerge from a class response map and then back-propagate them to map to highly informative regions of each object instance, such as instance boundaries.

    \item We implement the proposed method in popular CNNs, \eg, VGG16 and ResNet50, and show top performance on multiple benchmarks. To the best of our knowledge, we for the first time report results for the challenging image-level supervised instance segmentation task. 
\end{itemize}

%%%%%%%%% RELATED WORK
\section{Related Work}

\begin{figure}[!t]
    \begin{center}
        \includegraphics[width=1\linewidth]{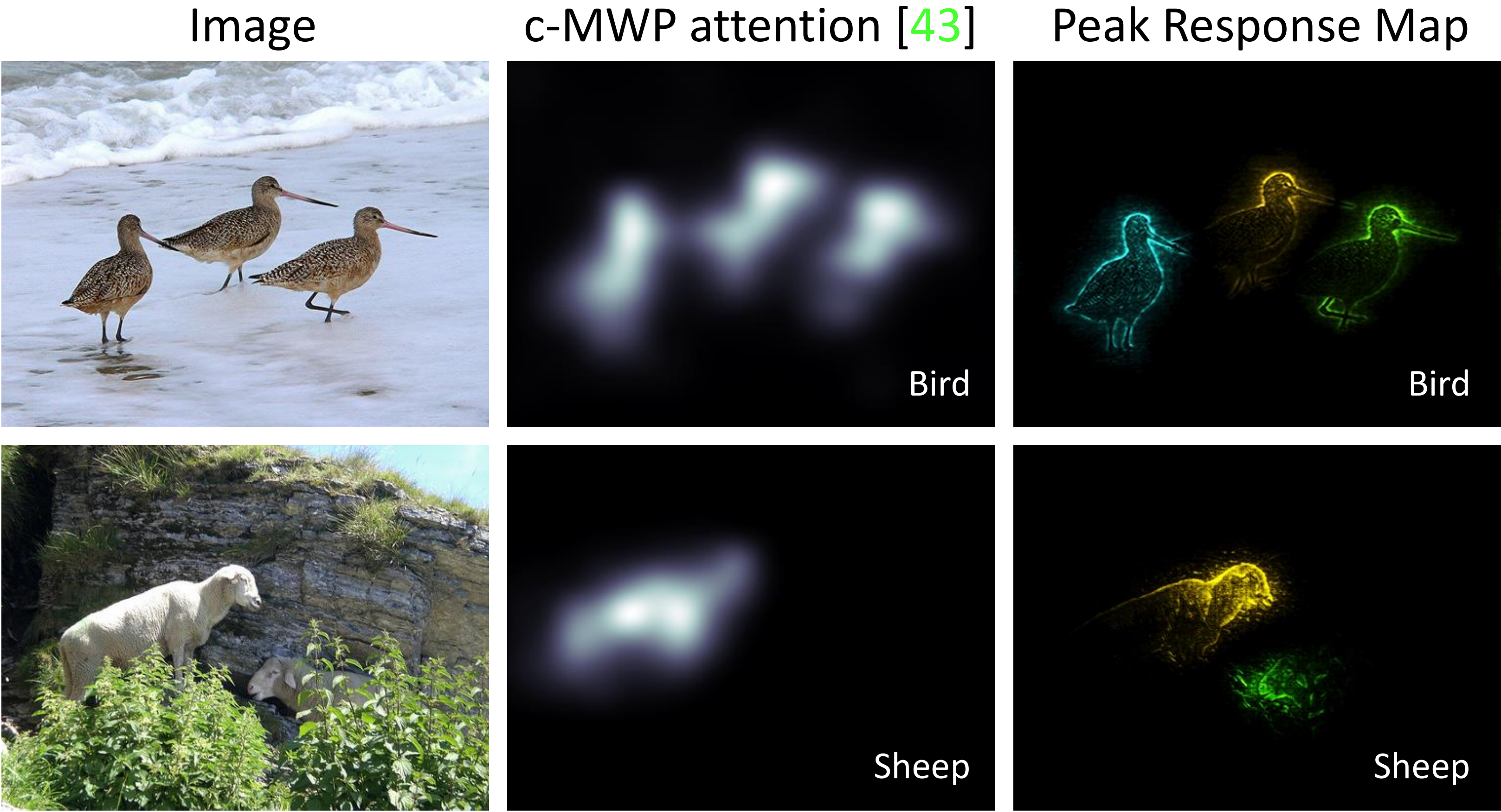}
    \end{center}
    \caption{
    Compared to existing weakly supervised methods which aim to obtain a saliency map (middle) for each \textbf{class}, the proposed approach extracts fine-detailed representation (right), including both explicit layouts and boundaries, for each \textbf{instance} (visualized with different colors).
    }
    \label{fig:compare}
    \vspace{-0.8em}
\end{figure}

\begin{figure*}[!htp]
    \centering
    \includegraphics[width=1\linewidth]{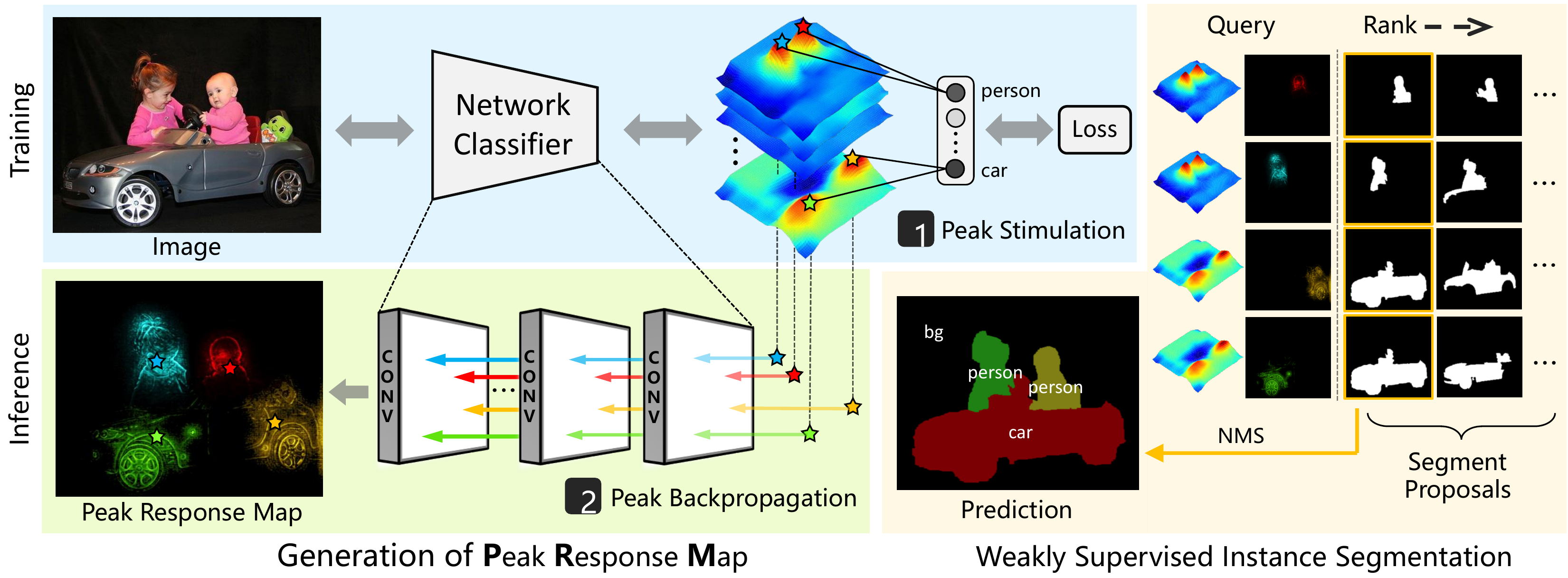}
    \caption{The generation and utilization of Peak Response Maps (PRMs). A stimulation procedure selectively activates strong visual cues residing inside each object into class peak responses. A back-propagation process further extracts fine details of each instance from the resulting peaks. Finally, class-aware cues, instance-aware cues, and object priors from proposals are considered together to predict instance masks. Best viewed in color.}
    \label{fig:peak-arch}
    \vspace{-0.2em}
\end{figure*}

\textbf{Weakly supervised semantic segmentation.}
Semantic segmentation approaches typically require dense annotations in the training phase. 
Given the inefficiency of pixel-level annotating,
previous efforts have explored various alternative weak annotations, \eg, points on instances \cite{bearman2016what}, object bounding boxes \cite{dai2015boxsup, george2015weakly}, scribbles \cite{lin2016scribblesup, xu2015learning}, and human selected foreground \cite{saleh2016built}.  
Although effective, these approaches require significant more human efforts than image-level supervised methods  \cite{deepak2015fully, pedro2015from, wei2017stc, kolesnikov2016seed, roy2017combining}.

Some works leverage object cues in an unsupervised manner. For examples, graphical models have been used to infer labels for segments \cite{zhang2015weakly, lai2016saliency}, yet their object localization capacity remains limited.  External localization network is therefore used to initialize object locations \cite{pinheiro2015weakly, kolesnikov2016seed, papandreou2015weakly}, and refining low-resolution CNN planes with pre-generated object segment proposal priors. Previous works usually involve time-consuming training strategies, \eg, repeatedly model learning \cite{wei2017object} or online proposal selection \cite{qi2016augmented, wan2018melm}.
In this work, instead, we use the standard classification networks to produce class-aware and instance-aware visual cues born with convolutional responses. 

\textbf{Instance segmentation.}
Compared with semantic segmentation that seeks to produce class-aware masks, instance segmentation requires to produce, at the same time, instance-aware region labels and fine-detailed segmentation masks and thus is much more challenging. Even with supervision from accurate pixel-level annotations, many instance segmentation approaches resort to additional constraints from precise object bounding boxes. 
The FCIS approach \cite{li2017fully} combines a segment proposal module \cite{dai2016instance} and an object detection system \cite{dai2016r}. Mask R-CNN \cite{he2017mask} fully leverages the precise object bounding boxes generated with a proposal network \cite{ren2015faster} to aid the prediction of object masks. 

With strong supervision from pixel-level GT masks, the above approaches have greatly boosted the performance of instance segmentation. However, the problem that how to perform instance segmentation under weak supervision remains open. Khoreva \etal \cite{khoreva2017simple} propose to obtain pseudo ground truth masks from bounding box supervision to alleviate labeling cost. In contrast, we leverage instance-aware visual cues naturally learned with classification networks; thus only image-level annotations are required for training.

\textbf{Object prior information.}
When accurate annotations are unavailable, visual recognition approaches leverage prior information typically to obtain additional visual cues. 
Object proposal methods that hypothesize object locations and extent are often used in weakly supervised object detection and segmentation to provide object priors. 
Selective Search \cite{uijlings2013selective} and Edge Boxes \cite{zitnick2014edge} use low-level features like color and edges as cues to produce object candidate windows. 
Multi-scale Combinatorial Grouping (MCG) \cite{jordi2015mcg} uses low-level contour information, \eg, Structured Edge \cite{dollar2015fast} or Ultrametric Contour Map \cite{kevis2017cob}, to extract object proposals, which contain fine-detailed object boundaries that is valuable to instance segmentation. 
In this paper, we perform instance mask extraction with the help of object priors from MCG proposals.

\textbf{Image-level supervised deep activation.}
With image-level supervision only, it is required to aggregate deep responses, \ie, feature maps, of CNNs into global class confidences so that image labels can be used for training.
Global max pooling (GMP) \cite{oquab2015object} chooses the most discriminative response for each class to generate classification confidence scores, but many other informative regions are discarded.
Global average pooling (GAP) \cite{zhou2016cnnlocalization} assigns equal importance to all responses, which makes it hard to differentiate foreground and background.
The log-sum-exponential (LSE) \cite{sun2016pronet} provides a smooth combination of GMP and GAP to constrain class-aware object regions.
Global rank max-min pooling (GRP) \cite{durand2017wildcat} selects a portion of high-scored pixels as positives and low-scored pixels as negatives to enhance discrimination capacity.

Existing approaches usually activate deep responses from a global perspective without considering local spatial relevance, which makes it hard to discriminate object instances in an image. 
Peaks in the convolution response imply a maximal local match between the learned filters and the informative receptive field.
In our method, the peak stimulation process aggregates responses from local maximums to enhance the network's localization ability.

Based on the deep responses, top-down attention methods are proposed to generate refined class saliency maps by exploring visual attention evidence \cite{cao2015look, zhang2016EB}. These class-aware and instance-agnostic cues can be used in semantic segmentation \cite{kolesnikov2016seed, roy2017combining} yet is insufficient for instance segmentation, Fig.~\ref{fig:compare}. In contrast, our methods provide fine-detailed instance-aware cues that are suitable for weakly supervised instance-level problems.

%%%%%%%%% METHOD
\section{Method}
In this section, we present an image-level supervised instance segmentation technique that utilizes class peak response.
CNN classifiers in the fully convolutional manner can produce class response maps, which specify classification confidence at each image location \cite{oquab2015object}. 
Based on our observation that local maximums, \ie, peaks, of class response maps typically correspond to strong visual cues residing inside an instance, we first design an process to stimulate peaks to emerge from a class response map in the network training phase.
During the inference phase, emerged peaks are back-propagated to generate maps that highlight informative regions for each object, referred to as Peak Response Maps (PRMs).
PRMs provide a fine-detailed separate representation for each instance, which are further exploited to retrieve instance masks from object segment proposals off-the-shelf, Fig.~\ref{fig:peak-arch}.

\begin{figure}[!htp]
    \centering
    \includegraphics[width=\linewidth]{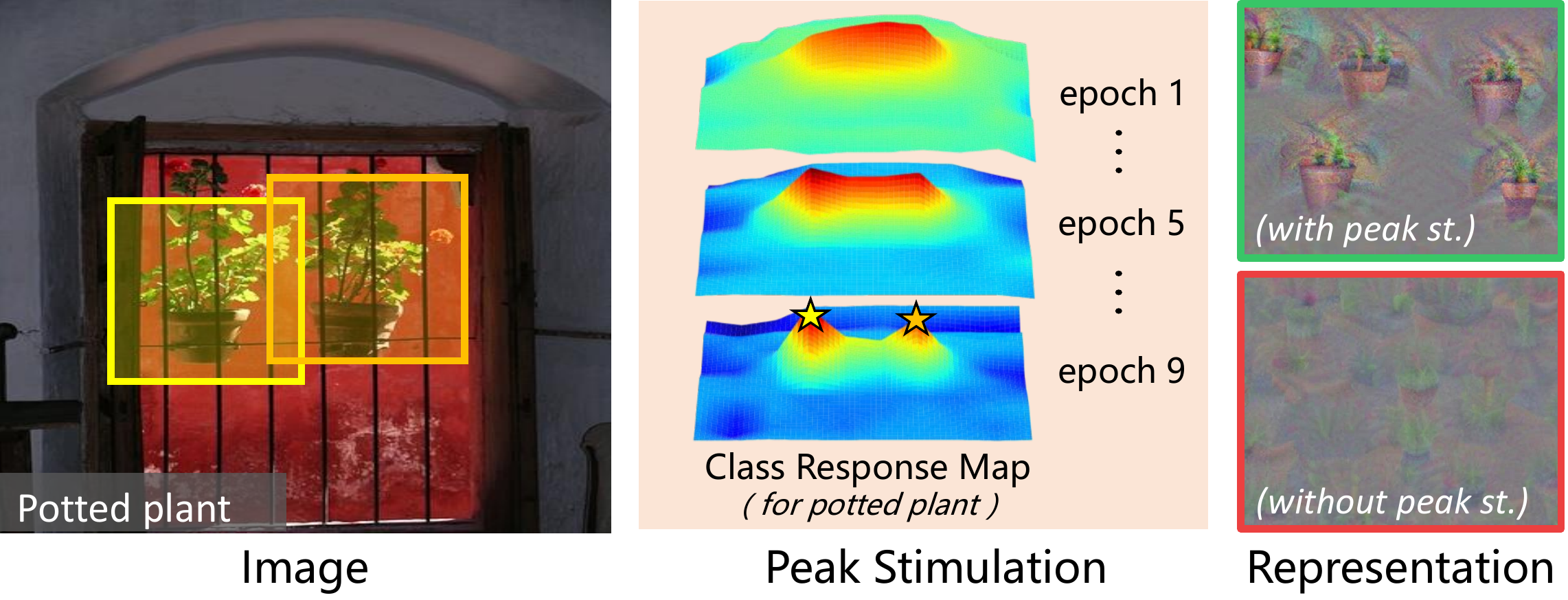}
    \caption{With peak stimulation, multiple instances can be better distinguished on the class response map (middle). The learned representations (right) are visualized by \textbf{activation maximization} \cite{erhan2009visualizing}. Best viewed in color.}
    \label{fig:peak-stimulation}
    \vspace{-0.8em}
\end{figure}
    
    \subsection{Fully Convolutional Architecture}
    By simply removing the global pooling layer and adapting fully connected layers to 1x1 convolution layers, modern CNN classifiers can be seamlessly converted to fully convolutional networks (FCNs) \cite{long2015fully} that naturally preserve spatial information throughout the forwarding. 
    The converted network outputs class response maps with a single forward pass; therefore are suitable for spatial predictions. In this work, networks are converted to FCN first. 
    
    \subsection{Peak Stimulation}
    \label{sec:peak-stimulation}
    To stimulate peaks to emerge from class response maps, we construct a peak stimulation layer, to be inserted after the top layer, Fig.~\ref{fig:peak-arch}.
    Consider a standard network, let $M \in \mathbb R^{C \times H \times W}$ denotes the class response maps of the top convolutional layer, where $C$ is the number of classes, and $H \times W$ denotes the spatial size of the response maps. Therefore, the input of the peak stimulation layer is $M$ and the output is class-wise confidence scores $s \in \mathbb R^{C}$. 
    Peaks of the $c$-th response map $M^c$ are defined to be the local maximums within a window size of $r$ \footnote{The region radius $r$ for peak finding is set to 3 in all our experiments.}, and the location of peaks are denoted as $P^c = \{(i_1, j_1), (i_2, j_2), ..., (i_{N^c}, j_{N^c})\}$, where $N^c$ is the number of valid peaks for the $c$-th class. During the forwarding pass, a sampling kernel $G^c \in \mathbb R^{H \times W}$ is generated for computing the classification confidence score of the $c$-th object category. Each kernel element at the location $(x,y)$ can be accessed with $G^c_{x,y}$. Without loss of generality, the kernel is formed as
    \begin{equation}
        G^c_{x,y} = \sum^{N^c}_{k=1}f(x - i_k, y - j_k),
        \label{eq:general-sampling}
    \end{equation}
    where $0 \leq x < H, 0 \leq y < W$, $(i_k, j_k)$ is the coordinate of the $k$-th peak, and $f$ is a sampling function. 
    In our settings, $f$ is a Dirac delta function for aggregating features from the peaks only; therefore the confidence score of the $c$-th category $s^c$ is then computed by the convolution between the class response map $M^c$ and sampling kernel $G^c$, as
    
    \begin{equation}
        s^c = M^c * G^c = \frac{1}{N^c}\sum^{N^c}_{k=1}M^c_{i_k,j_k}.
        \label{eq:peak-pooling-forward}
    \end{equation}
    
    It can be seen from Eq.~\ref{eq:peak-pooling-forward} that the network uses peaks only to make the final decision; naturally, during the backward pass, the gradient is apportioned by $G^c$ to all the peak locations, as
    \begin{equation}
        \delta^c = \frac{1}{N^c} \cdot \frac{\partial{L}}{\partial{s^c}} \cdot G^c,
        \label{eq:peak-pooling-backward}
    \end{equation}
    where $\delta^c$ is the gradient for the $c$-th channel of the top convolutional layer and $L$ is the classification loss. 
    
    From the perspective of model learning, the class response maps are computed by the dense sampling of all receptive fields (RFs), in which most of RFs are negative samples that do not contain valid instances. Eq.~\ref{eq:peak-pooling-backward} indicates that in contrast to conventional networks which unconditionally learn from the extreme foreground-background imbalance set, peak stimulation forces the learning on a sparse set of informative RFs (potential positives and hard negatives) estimated via class peak responses, thus prevents the vast number of easy negatives from overwhelming the learned representation during training, Fig.~\ref{fig:peak-stimulation} (right).
    
    \begin{figure*}[!htp]
        \centering
        \includegraphics[width=1\linewidth]{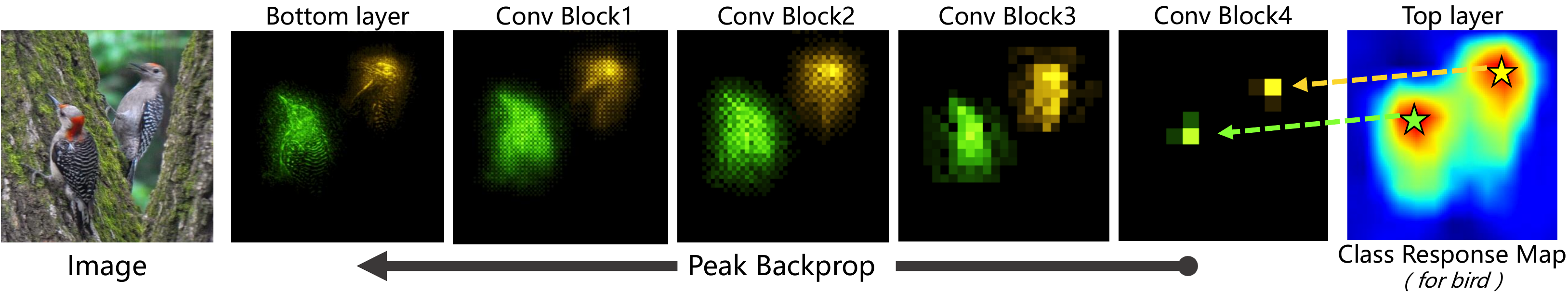}
        \caption{Peak back-propagation process maps class peak responses to fine detailed visual cues residing inside each object, \ie, Peak Response Maps (PRMs), enabling the instance-level masks to be extracted. Best viewed in color.}
        \label{fig:peak-backprop}
    \end{figure*}
    
    \subsection{Peak Back-propagation}
    \label{sec:peak-backprop}
    
    We propose a probability back-propagation process for peaks to further generate the fine-detailed and instance-aware representation, \ie, Peak Response Map. In contrast to previous top-down attention models \cite{zhang2016EB, tsotsos1995modeling}, which seek the most relevant neurons of an output category to generate class-aware attention maps, our formulation explicitly considers the receptive field and can extract instance-aware visual cues from the specific spatial locations, \ie, class peak responses. The peak back-propagation can be interpreted as a procedure that a walker starts from the peak (top layer) and walk randomly to the bottom layer. The top-down relevance of each location in the bottom layer is then formulated as its probability of being visited by the walker.
    
    Consider a convolution layer that has a single filter $W \in \mathbb R^{kH \times kW}$ for mathematical simplification, the input and output feature maps are denoted as $U$ and $V$, where each spatial locations can be accessed by $U_{ij}$ and $V_{pq}$ respectively.
    The visiting probability $P(U_{ij})$ can be obtained by $P(V_{pq})$ and the transition probability between two maps, as
    \begin{equation}
        P(U_{ij}) = \sum_{p=i-\frac{kH}{2}}^{i+\frac{kH}{2}}\sum_{q=j-\frac{kW}{2}}^{j+\frac{kW}{2}}P(U_{ij}|V_{pq}) \times P(V_{pq}),
        \label{eq:backprop}
    \end{equation}    
    where the transition probability is defined as    
    \begin{equation}
        P(U_{ij}|V_{pq}) = Z_{pq} \times \hat{U}_{ij}W^+_{(i-p)(j-q)}.
        \label{eq:backprop-transfer}
    \end{equation}
    $\hat{U}_{ij}$ is the bottom-up activation (computed in the forward pass) at the location $(i,j)$ of $U$, $W^+ = ReLU(W)$, which discards negative connections, and $Z_{pq}$ is a normalization factor to guarantee $\sum_{p,q} P(U_{ij}|V_{pq}) = 1$. Note in most modern CNNs that adopt ReLU as transfer function, negative weights have no positive effects in enhancing the output response, thus are excluded from propagation.
    
    Other commonly used intermediate layers, \eg, the average pooling and max-pooling layers, are regarded as the same type of layers that perform an affine transform of the input \cite{zhang2016EB}; thus the corresponding back-propagation can be modeled in the same way of convolution layers. 
    
    With the probability propagation defined by Eq.~\ref{eq:backprop} and Eq.~\ref{eq:backprop-transfer}, we can localize most relevant spatial locations for each class peak response in a top-down fashion, to generate fine-detailed instance-aware visual cues, referred to as Peak Response Map, Fig.~\ref{fig:peak-backprop}.
    
    \subsection{Weakly Supervised Instance Segmentation}
    \label{sect:method}
    We further leverage the instance-aware cues of PRMs to perform challenging instance segmentation tasks.
    Specifically, we propose a simple yet effective strategy to predict mask for each object instance by combining instance-aware cues from PRMs, class-aware cues from class response maps, and spatial continuity priors from object proposals off-the-shelf \cite{uijlings2013selective, jordi2015mcg, kevis2017cob}.
    
    We retrieve instance segmentation masks from a proposal gallery, Fig.~\ref{fig:peak-arch}, with the metric,
    \begin{equation}
        Score =   \underbrace{\vphantom{\beta \cdot Q} \alpha \cdot R * S}_{\text{instance-aware}}
                + \underbrace{\vphantom{\beta \cdot Q} R * \hat{S}}_{\text{boundary-aware}}
                - \underbrace{\vphantom{\beta \cdot Q} \beta \cdot Q * S}_{\text{class-aware}},
        \label{eq:metric}
        \vspace{-0.2em}
    \end{equation}
    where $R$ is the PRM corresponds to a class peak response, $\hat{S}$ is the contour mask of the proposal $S$ computed by morphological gradient, and $Q$ is the background mask obtained by the class response map and a bias (based on the mean value of the map). The class independent free parameters $\alpha$ and $\beta$ are selected on the validation set.
    
    In Eq.~\ref{eq:metric}, the instance-aware term encourages proposal to maximize the overlap with PRM, while the boundary-aware term leverages the fine-detailed boundary information within the PRM to select proposal with a similar shape. Furthermore, the class-aware term uses class response map to suppress class-irrelevant regions. The effects of three terms are ablation studied in Sec.~\ref{sect:ws-inst-seg}.
    
    The overall algorithm for weakly supervised instance segmentation is specified in Alg.~\ref{alg:A1}.
    
    {
    \setlength{\textfloatsep}{2pt}
    \renewcommand{\algorithmicrequire}{\textbf{Input:}} 
    \renewcommand{\algorithmicensure}{\textbf{Output:}}
    \begin{algorithm}[!t]
      \begin{algorithmic}[1]
        \Require{A test image $I$, segment proposals $\mathcal S$, and a network trained with peak stimulation.}
        \Ensure{Instance segmentation prediction set $\mathcal A$}
        \State Initialize instance prediction set $\mathcal A=\emptyset$;
        \State Forward $I$ to get class response maps $M$;
        \For{map $M^k$ of the $k$-th class in $M$}
            \State Detect peaks $P_i^k$ and add to $\mathcal P$, Sec.~\ref{sec:peak-stimulation};
        \EndFor
        
        \For{peak $P_i^k$ in $\mathcal P$}
            \State Peak backprop at $P_i^k$ to get PRM $R$, Sec.~\ref{sec:peak-backprop};
            \For{proposal $S_j$ in $\mathcal S$}
                \State Compute score using $R$ and $M^k$, Eq.~\ref{eq:metric};
            \EndFor
            \State Add top-ranked proposal and label $(S_*, k)$ to $\mathcal A$;
        \EndFor
        \State Do Non-Maximum Suppression (NMS) over $\mathcal A$. 
      \end{algorithmic}
        \caption{Segment Instances via Class Peak Response}
        \label{alg:A1}
    \end{algorithm}
    }

%%%%%%%%% EXPERIMENT
\section{Experiment}
We implement the proposed method using state-of-the-art CNN architectures, including VGG16 and ResNet50, and evaluate it on several benchmarks.
In Sec.~\ref{sect:peak-response-quality}, we perform a detailed analysis of the peak stimulation and back-propagation process, to show that the proposed technique can generate accurate object localization and high-quality instance-aware cues.
In Sec.~\ref{sect:ws-semantic-seg}, on weakly supervised semantic segmentation, the ability of PRMs to extract class-aware masks with the help of segment proposals is shown.
In Sec.~\ref{sect:ws-inst-seg}, we for the first time report results for challenging image-level supervised instance segmentation. Ablation study and upper bound analysis are further performed to demonstrate the effectiveness and potential of our method.

    \subsection{Peak Response Analysis}
    \label{sect:peak-response-quality}

    \textbf{Pointwise localization.}
    A pointwise object localization metric \cite{oquab2015object} is used to evaluate the localization ability of class peak responses and effectiveness of peak stimulation. We first upsample the class response maps to the size of the image via bilinear interpolation. For each predicted class, if the coordinate of the maximum class peak response falls into a ground truth bounding box of the same category, we count a true positive.

    We fine-tune ResNet50 equipped with/without peak stimulation on the training set of PASCAL VOC 2012 \cite{everingham2015the} as well as MS COCO 2014 \cite{lin2014microsoft}, and report performances on the validation set, Tab.~\ref{tab:point-loc-map-voc12}. The results show that class peak responses correspond to visual cues of objects and can be used to localize objects. Our full approach shows top performance against state-of-the-arts and outperforms the baseline (w/o stimulation) by a large margin, which indicates the stimulation process can lead the network to discover better visual cues correspond to valid instances.

    \begin{table}[!t]
      \begin{center}
        \begin{tabular}{l|cc}
            \hline
            Method  & VOC 2012 & MS COCO \\ \hline\hline
            DeepMIL \cite{oquab2015object} & 74.5 & 41.2  \\
            WSLoc \cite{bency2016weakly}  & 79.7 & 49.2   \\ 
            WILDCAT \cite{durand2017wildcat} & 82.9 & 53.5 \\
            SPN \cite{zhu2017soft} & 82.9 &  55.3 \\ \hline
            Ours (w/o Peak Stimulation) & 81.5 &  53.1 \\
            Ours (full approach) & \textbf{85.5} & \textbf{57.5} \\ \hline 
        \end{tabular}
      \end{center}
      \caption{Mean Average Precision (mAP\%) of pointwise localization on VOC2012 and COCO2014 val. set.}
      \label{tab:point-loc-map-voc12}
      \vspace{-0.8em}
    \end{table}

    \textbf{Quality of peak response maps.}
    To evaluate the quality of extracted instance-aware cues, we measure the correlation between a Peak Response Map (PRM) $R$ and a GT mask $G$ with $\frac{\sum R \odot G}{\sum R}$, which indicates the ability of the PRM to discover visual cues residing inside the instance. For each PRM, we define its score to be the largest correlation with GT masks of the same class. Thus, a score of 0 indicates that the corresponding PRM does not locate any valid object region, while a score of 1 implies the PRM perfectly distinguishes the visual cues of an instance from the background. PRMs with a score higher than $0.5$ are considered as true positives.
    On VOC 2012, we use classification data to train ResNet50 equipped with response aggregation strategies from different methods, and evaluate the quality of resulting PRMs on the validation set of the segmentation data in terms of mAP, Tab.~\ref{tab:peak-energy-map}. Peak stimulation forces networks to learn an explicit representation from informative receptive fields; thus obtaining higher quality of PRMs.

\begin{figure}[!t]
        \begin{center}
            \includegraphics[width=1\linewidth]{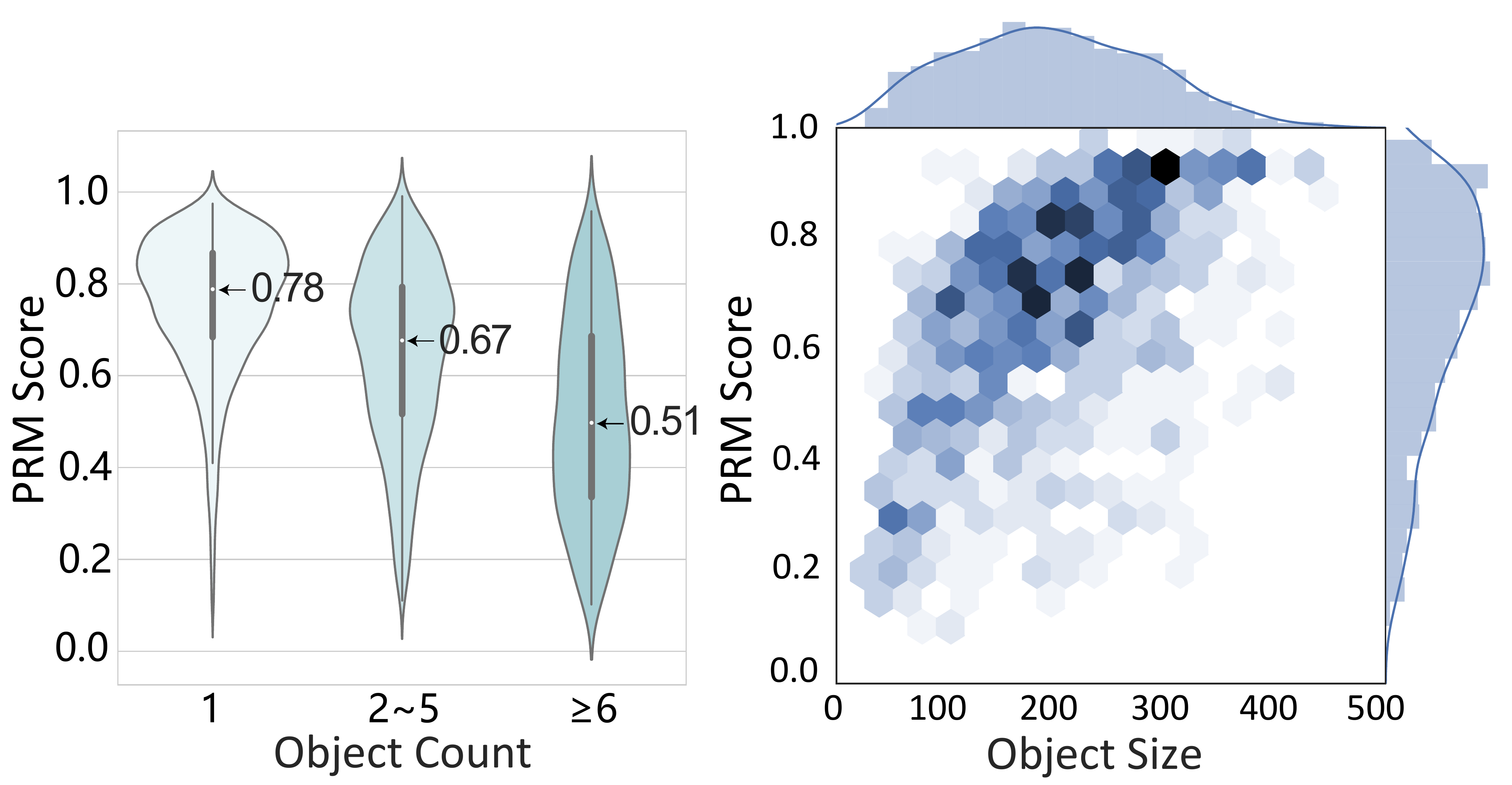}
        \end{center}
        \caption{Statistical analysis of the effect of the number and size of objects on the quality of Peak Response Maps.}
        \label{fig:peak-energy}
        \vspace{-0.2em}
    \end{figure}

    We perform statistical analysis on the relationship between the PRM quality and the crowding level of images, Fig.~\ref{fig:peak-energy} (left). On average, the energy of PRMs that falls into an instance reaches $78\%$ for images with a single object, and $67\%$ for images with 2-5 objects. Surprisingly, even for crowded scenes with more than six objects, the instances collect more energy than the background on average, which shows that the instance-aware visual cues from PRMs are of high quality.
    We further analyze the impact of object size, Fig.~\ref{fig:peak-energy} (right), and results show that PRMs can localize fine-detailed evidence from common size objects.

    \begin{table}[!t]
        \begin{center}
            \begin{tabular}{c|c|c}
                \hline
                Method  & Response Aggregation Strategy & mAP \\ \hline \hline
                CAM \cite{zhou2016cnnlocalization} & Global Average Pooling & 55.7 \\
                DeepMIL \cite{oquab2015object} & Global Max Pooling & 60.9 \\
                WILDCAT \cite{durand2017wildcat} & Global Max-Min Pooling & 62.4 \\ \hline
                PRM (Ours) & Peak Stimulation & \textbf{64.0} \\
                \hline
            \end{tabular}
        \end{center}
        \caption{Comparison of the effect of different response aggregation strategies on the quality of Peak Response Maps.}
        \label{tab:peak-energy-map}
        \vspace{-0.8em}
    \end{table}

    \subsection{Weakly Supervised Semantic Segmentation}
    \label{sect:ws-semantic-seg}
    Experiments above shows that the PRMs correspond to accurate instance ``seeds" while another challenging thing is to expand each seed into full object segmentation. We evaluate the ResNet50 model equipped with peak stimulation on the weakly supervised semantic segmentation task, which requires assigning objects from the same categories as the same segmentation labels. On the validation set of VOC 2012 segmentation data, We merge the instance segmentation masks of the same class to produce semantic segmentation predictions. The performance is measured regarding pixel intersection-over-union averaged across 20+1 classes (20 object categories and background).

\begin{figure}[!t]
        \begin{center}
            \includegraphics[width=0.9\linewidth, height=0.62\linewidth]{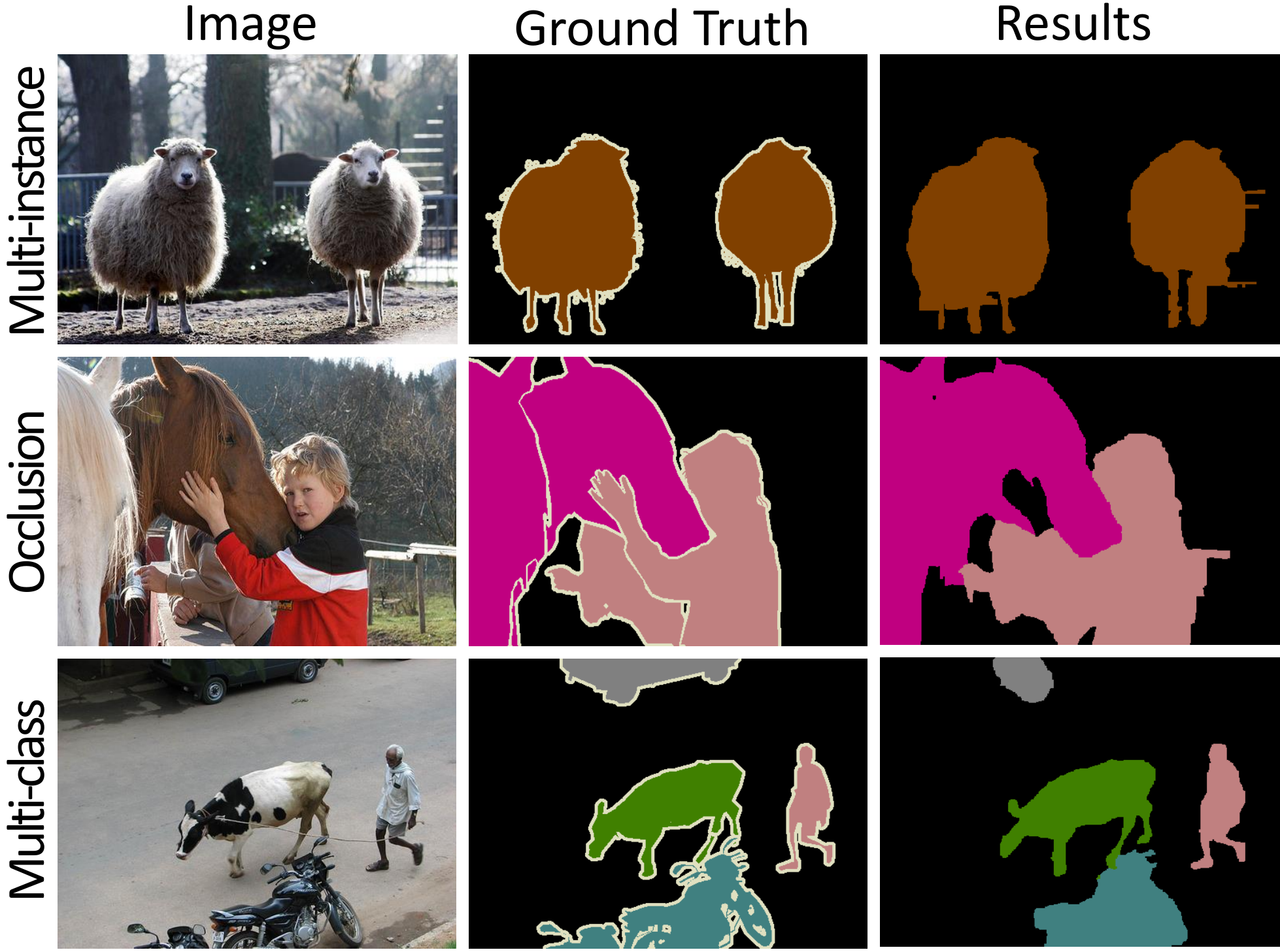}
        \end{center}
          \caption{Examples of predicted semantic segmentations. Different colors indicate different classes.}
        \label{fig:se-seg-sample}
        \vspace{-0.8em}
        \end{figure}

    Instead of using time-consuming training strategies \cite{roy2017combining}, or additional supervisions \cite{bearman2016what, saleh2016built}, our method trains models
    using image-level labels and standard classification settings, and reports competitive results, on weakly supervised semantic segmentation without CRF post-processing, Tab.~\ref{tab:se-seg-voc12-compare-supervision}. Fig.~\ref{fig:se-seg-sample} shows examples of predictions in different scenarios. 

        \begin{table}[!t]
            \begin{center}
            \resizebox{0.98\linewidth}{!}{
                \begin{tabular}{l|c|c}
                \hline
                Method &  mIoU & Comments\\ \hline \hline
                MIL+ILP+SP-seg $^\dag$ \cite{pinheiro2015weakly} & 42.0 & Object segment proposals \\ 
                WILDCAT $^\dag$ \cite{durand2017wildcat}  & 43.7 & CRF post-processing \\
                SEC \cite{kolesnikov2016seed} & 50.7 & CRF as boundary loss \\
                Check mask \cite{saleh2016built} & 51.5 & CRF \& Human in the loop \\
                Combining \cite{roy2017combining} & 52.8 & CRF as RNN \\ \hline
                PRM (Ours) $^\dag$ & \textbf{53.4} & Object segment proposals \\ \hline
                \end{tabular}
                }
            \end{center}
            \caption{Weakly supervised semantic segmentation results on VOC 2012 val. set in terms of the mean IoU (\%). Mark $\dag$ indicates methods that introduce negligible training costs.}
            \label{tab:se-seg-voc12-compare-supervision}
            \vspace{-0.8em}
        \end{table}
        
    \subsection{Weakly Supervised Instance Segmentation}
    \label{sect:ws-inst-seg}

    With the proposed technique, we perform instance segmentation on the PASCAL VOC 2012 segmentation set with ResNet50 and VGG16 models trained on the classification set. To the best of our knowledge, this is the first work reporting results for image-level supervised instance segmentation. We construct several baselines based on object bounding boxes obtained from ground truth and weakly supervised localization methods \cite{wan2018melm, zhou2016cnnlocalization, zhu2017soft}, Tab.~\ref{tab:inst-seg-voc12-map}.
    With the localized bounding boxes, we set three reasonable mask extraction strategies: (1) Rect. Simply filling in the object boxes with instance labels, (2) Ellipse. Fitting a maximum ellipse inside each box, and (3) MCG. Retrieving an MCG segment proposal of maximum IoU with the bounding box.

    \textbf{Numerical results.} 
    The instance segmentation is evaluated with the $mAP^r$ at IoU threshold 0.25, 0.5 and 0.75, and the Average Best Overlap (ABO) \cite{pont2015boosting} metric is also employed for evaluation to give a different perspective. 
    Tab.~\ref{tab:inst-seg-voc12-map} shows that our approach significantly outperforms weakly supervised localization techniques that use the same setting, \ie, using image-level labels only for model training.
    The performance improvement at lower IoU thresholds, \eg, 0.25 and 0.5, shows the effectiveness of peak stimulation for object location, while the improvement at higher IoU threshold, \eg, 0.75, indicates the validity of peak back-propagation for capturing fine-detailed instance cues.

    Compare with the latest state-of-the-art MELM \cite{wan2018melm}, which is trained with multi-scale augmentation, online proposal selection, and a specially designed loss, our method is simple yet effective and shows a competitive performance.

    \begin{table}[!t]
        \begin{center}
        \resizebox{\linewidth}{!}{
        \begin{tabular}{c|c|cccc}
        \hline
        \multicolumn{2}{c|}{Method}
            & $mAP^r_{0.25}$ & $mAP^r_{0.5}$ & $mAP^r_{0.75}$ & ABO \\ \hline \hline
        \multirow{3}{*}{\begin{tabular}[c]{@{}c@{}}Ground\\ Truth\end{tabular}}                                      
            & Rect.     & 78.3 & 30.2 & 4.5 & 47.4 \\ \cline{2-6} 
            & Ellipse   & 81.6 & 41.1 & 6.6 & 51.9 \\ \cline{2-6}   
            & MCG       & 69.7 & 38.0 & 12.3 & 53.3 \\ \hline
        \multicolumn{6}{c}{\cellcolor{lightgray}Training requires image-level labels and object proposals} \\ \hline
        \multirow{3}{*}{MELM \cite{wan2018melm}}                                                           
            & Rect.     & 36.0 & 14.6 & 1.9 & 26.4 \\ \cline{2-6} 
            & Ellipse   & 36.8 & 19.3 & 2.4 & 27.5 \\ \cline{2-6} 
            & MCG       & 36.9 & 22.9 & 8.4 & 32.9 \\ \hline 
        \multicolumn{6}{c}{\cellcolor{lightgray}Training requires only image-level labels} \\ \hline
        \multirow{3}{*}{CAM \cite{zhou2016cnnlocalization}}
            & Rect.     & 18.7 & 2.5 & 0.1 & 18.9 \\ \cline{2-6} 
            & Ellipse   & 22.8 & 3.9 & 0.1 & 20.8 \\ \cline{2-6} 
            & MCG       & 20.4 & 7.8 & 2.5 & 23.0 \\ \hline 
        \multirow{3}{*}{SPN \cite{zhu2017soft}}                                                            
            & Rect.     & 29.2 & 5.2 & 0.3 & 23.0 \\ \cline{2-6} 
            & Ellipse   & 32.0 & 6.1 & 0.3 & 24.0 \\ \cline{2-6} 
            & MCG       & 26.4 & 12.7 & 4.4 & 27.1 \\ \hline
        \multicolumn{2}{c|}{PRM (Ours)} 
            & \textbf{44.3} & \textbf{26.8} & \textbf{9.0} & \textbf{37.6} \\ \hline
        \end{tabular}
        }
        \end{center}
        
        \caption{Weakly supervised instance segmentation results on the PASCAL VOC 2012 val. set in terms of mean average precision (mAP\%) and Average Best Overlap (ABO).}
        \label{tab:inst-seg-voc12-map}
    \end{table}

    \begin{table}[!t]
        \begin{center}
            \resizebox{\linewidth}{!}{
            \begin{tabular}{c|ccccc|cc}
                \hline
                \multicolumn{1}{l|}{} & \multicolumn{5}{c|}{ResNet50} & \multicolumn{2}{c}{VGG16} \\ \hline \hline
                Peak Stimulation & & \checkmark & \checkmark & \checkmark & \checkmark & & \checkmark      \\ 
                Instance-aware term & \checkmark & & \checkmark & \checkmark & \checkmark & \checkmark & \checkmark         \\ 
                Class-aware term & \checkmark & \checkmark &  & \checkmark & \checkmark & \checkmark & \checkmark  \\ 
                Boundary-aware term & \checkmark & \checkmark &\checkmark &  & \checkmark & \checkmark & \checkmark \\ \hline 
                $mAP^r_{0.5}$ & 22.8 & 13.3 & 16.5 & 24.3 & \textbf{26.8} & 11.9 & 22.0 \\ \hline 
            \end{tabular}
            }
        \end{center}
        \caption{Ablation study on the PASCAL VOC2012 val. set based on different network backbones.}
        \label{tab:inst-seg-ablation}
        \vspace{-1em}
    \end{table}

    \begin{figure*}[!t]
        \begin{center}
            \includegraphics[width=0.93\linewidth, height=0.65\linewidth]{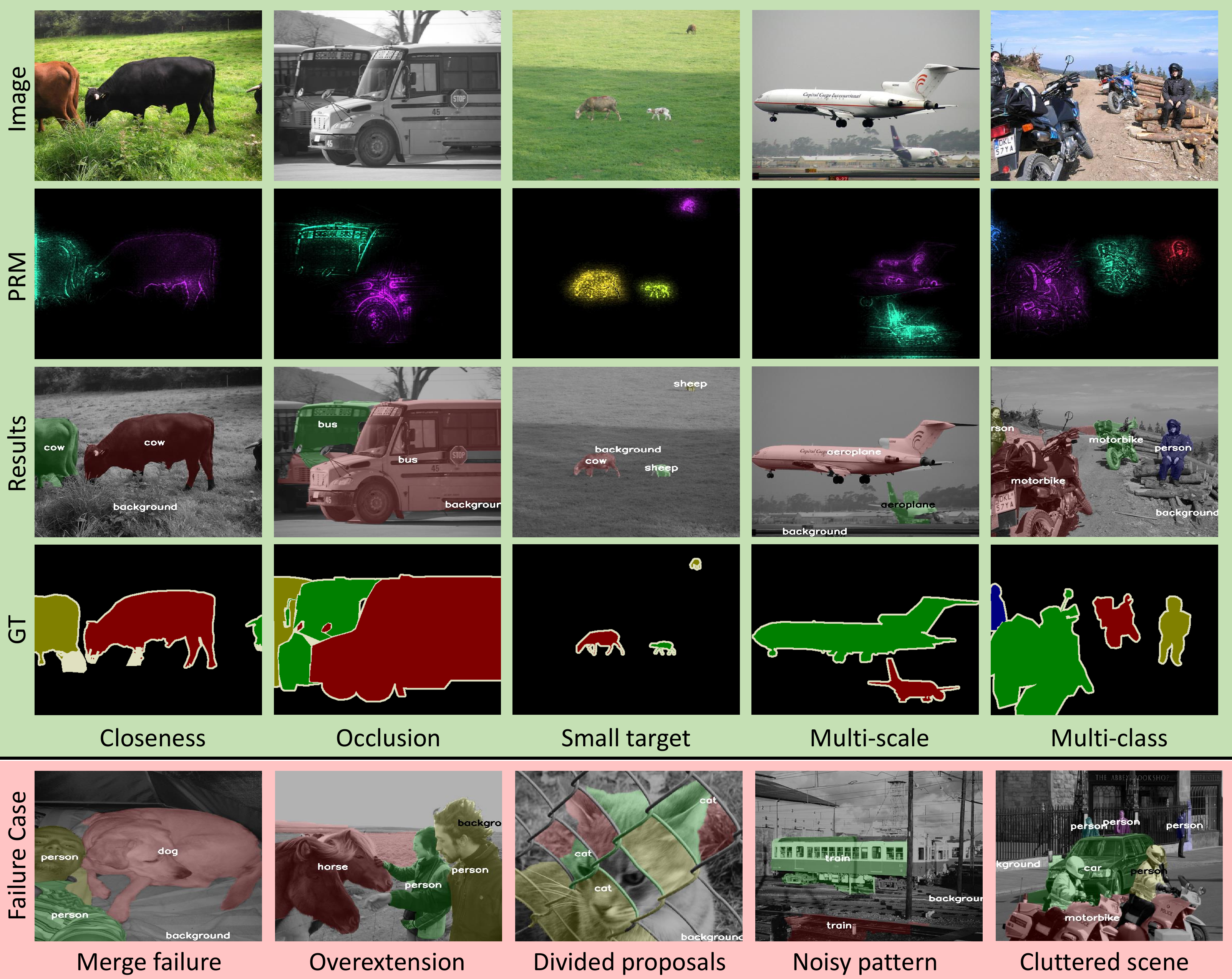}
        \end{center}
        \caption{Instance segmentation examples on the PASCAL VOC 2012 val. set. It can be seen that the Peak Response Maps (second row) incorporate fine-detailed instance-aware information, which can be exploited to produce instance-level masks (third row). The last row shows typical failure cases. Best viewed in color.}
        \label{fig:ins-seg-samples}
    \end{figure*}
        
    \textbf{Ablation study.}
    To investigate the contribution of peak stimulation as well as each term in our proposal retrieval metric, we perform instance segmentation based on different backbones in which different factors were omitted.
    The results are presented in Tab.\ \ref{tab:inst-seg-ablation}. From the ablation study, we can draw the following conclusions: 1). Peak stimulation process, which stimulates peaks during network training, is crucial to the instance segmentation performance of our method. 2). The $mAP^r_{0.5}$ dramatically drops from 26.8\% to 13.3\% when omitting the instance-aware term, which demonstrates the effectiveness of the well-isolated instance-aware representation generated by our method.
    3). Boundary-aware term significantly improves the performance by 2.5\% shows our method does extract fine-detailed boundary information of instances.
    4). Class-aware cues depress class-irrelevant regions; thus substantially improve the instance segmentation performance of our method.

    \textbf{Qualitative results.}
    In Fig.~\ref{fig:ins-seg-samples}, we illustrate some instance segmentation examples including successful cases and typical failure cases. 
    It can be seen that our approach can produce high quality visual cues and obtain decent instance segmentation results in many challenging scenarios.
    In the first and second columns, it can distinguish instances when they are closed or occluded with each other. Examples in the third and fourth columns show that it performs well with objects from different scales.
    In the fifth column, objects from different class are well segmented, which shows that the proposed method can extract both class-discriminative and instance-aware visual cues from classification networks.
    As is typical for weakly-supervised systems, PRMs can be misled by noisy co-occurrence patterns and sometimes have problems telling the difference between object parts and multiple objects. We address this problem with a proposal retrieval step; nevertheless, the performance remains limited by proposal quality.

    \begin{table}[!tp]
        \begin{center}
            \resizebox{1\linewidth}{!}{%
            \begin{tabular}{|l|cc|cc|}
            \hline
            Proposal gallery & GT mask & GT bbox  & SPN \cite{zhu2017soft} & PRM (Ours) \\ \hline \hline
            MCG & 26.0   & 12.3     & 4.4   & 9.0\\ \hline
            MCG + GT mask & 100.0   & 29.2  & 10.4 & 26.9\\ \hline
            GT mask & 100.0   & 93.0     & 50.0  & 73.3\\ \hline
            \end{tabular}%
            }
        \end{center}
        \caption{Comparison of instance segmentation results ($mAP^r_{0.75}$) on the PASCAL VOC 2012 val. set.}
        \label{tab:upper-bound}
        \vspace{-1.2em}
    \end{table}

    \textbf{Upper bound analysis.}
    To explore the upper bound of our method, we construct different proposal galleries, Tab.~\ref{tab:upper-bound}.
    First, we mix GT masks into MCG proposals to get a gallery with 100\% recall, and the results show that the capability of our method (image-level supervised) to retrieve proposals is comparable to GT bbox (26.9\% vs. 29.2\%). 
    Next, we use GT masks as a perfect proposal gallery (note that GT bbox still fails in highly occlusion cases) to evaluate the instance localization ability of PRMs. Our result further boosts to 73.3\% and outperforms SPN by a large margin, demonstrating the potential of the proposed technique on video/RGB-D applications where rich information can be exploited to generate proposals of high quality.

%%%%%%%%% CONCLUSIONS
\section{Conclusions}
In this paper, we propose a simple yet effective technique to enable classification networks for instance mask extraction.
Based on class peak responses, the peak stimulation shows effective to reinforce object localization, while the peak back-propagation extracts fine-detailed visual cues for each instance. 
We show top results for pointwise localization as well as weakly supervised semantic segmentation and, to the best of our knowledge, for the first time report results for image-level supervised instance segmentation.
The underlying fact is that instance-aware cues are naturally learned by convolutional filters and encoded in hierarchical response maps. To discover these cues provides fresh insights for weakly supervised instance-level problems.

%%%%%%%%% ACKNOWLEDGEMENTS
\section*{Acknowledgements}
The authors are very grateful for support by the NSFC grant 61771447 / 61671427, BMSTC, and NSF.

\clearpage

{\small
\bibliographystyle{ieee}
\bibliography{egbib}
}

\end{document}